\documentclass[11pt,a4paper]{article}
\usepackage[hyperref]{acl2019}
\usepackage{times}
\usepackage{latexsym}
\usepackage[T1]{fontenc}   
\usepackage{txfonts}
\usepackage{mathptmx}
\usepackage{color}
\usepackage{booktabs}
\usepackage{textcomp}
\usepackage{url}
\usepackage{balance}       
\usepackage{graphicx}      
\usepackage{microtype}        

\aclfinalcopy 

\newcommand\emoji[1]{\includegraphics[width=1em,trim={0 0 0 0},clip]{emoji_images/#1.png}}

\title{Federated Learning for Emoji Prediction in a Mobile Keyboard}

\author{Swaroop Ramaswamy\qquad Rajiv Mathews\qquad Kanishka Rao\qquad Fran{\c{c}}oise Beaufays \\
Google LLC \\
  Mountain View, CA, U.S.A. \\
  {\tt \{swaroopram, mathews, kanishkarao, fsb\}@google.com}}

\date{}

\begin{document}
\maketitle
\begin{abstract}
We show that a word-level recurrent neural network can predict emoji from text
typed on a mobile keyboard. We demonstrate the usefulness of transfer learning
for predicting emoji by pretraining the model using a language modeling task.
We also propose mechanisms to trigger emoji and tune the diversity of
candidates. The model is trained using a distributed on-device learning
framework called federated learning. The federated model is shown to achieve
better performance than a server-trained model. This work demonstrates the
feasibility of using federated learning to train production-quality models for
natural language understanding tasks while keeping users' data on their devices.

\end{abstract}

\section{Introduction}
Emoji have become an important mode of expression on smartphones as users
increasingly use them to communicate on social media and chat applications.
Easily accessible emoji suggestions have therefore become an important feature
for mobile keyboards.

\begin{figure}
  \includegraphics[width=\columnwidth]{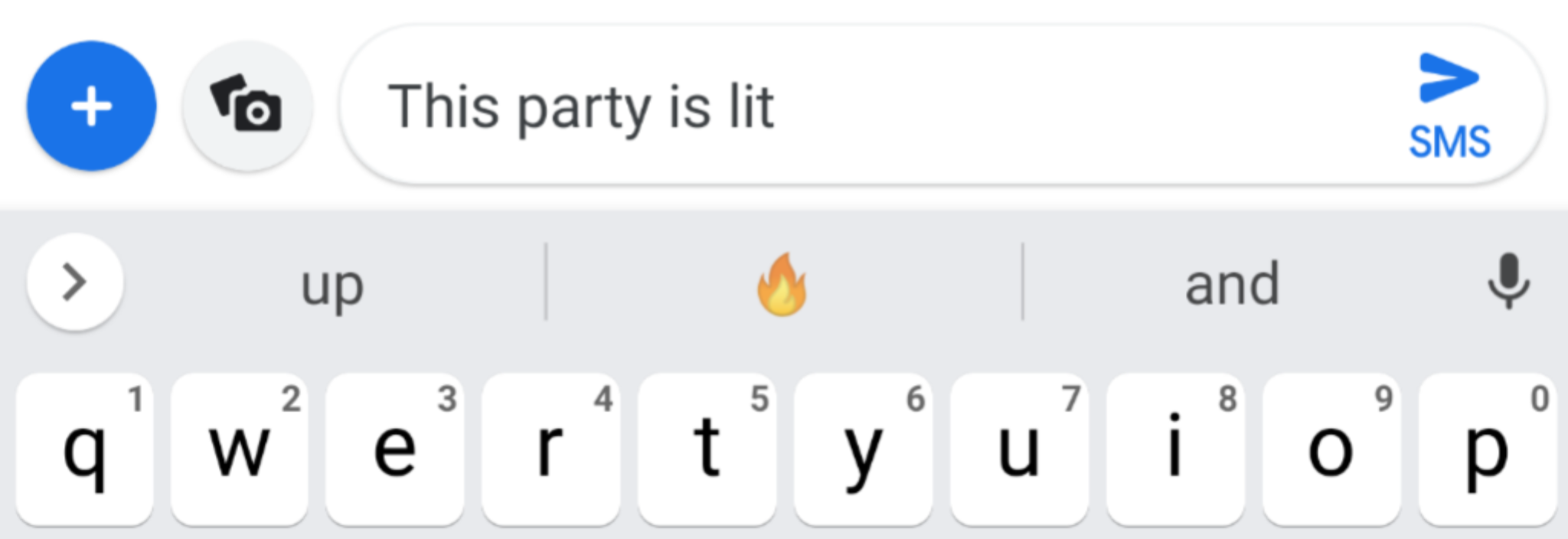}
  \caption{Emoji predictions in Gboard. Based on the context ``This party is lit'',
           Gboard predicts both emoji and words.}
  ~\label{fig:screenshot}
\end{figure}

Gboard
is a mobile keyboard with more than 1 billion installs and support for over 600
language varieties. With this work, we provide a mechanism by which Gboard offers\
emoji as predictions based on the text previously typed, as shown in
Figure \ref{fig:screenshot}.

Mobile devices are constrained by both memory and CPU. Low-latency is also
required, since users typically expect a keyboard response within 20
ms  of an input event \cite{gboardfst}.

A unidirectional recurrent neural network architecture (RNN) is used in this
work. Since forward RNNs only include dependencies backwards in time, the model state can be cached at each timestep during
inference to reduce latency.

\section{Federated Learning}
Federated Learning (FL)~\cite{bonawitz2019towards} is a new computation paradigm
in which data is kept on users' devices and never collected centrally.
Instead, minimal and focused model updates are transmitted to the server.
This allows us to train models while keeping users' data on their devices. FL can be combined with other privacy-preserving techniques like secure multi-party
computation~\cite{Bonawitz2017} and differential
privacy~\cite{McMahan2017LearningDP, agarwal2018cpsgd, abadi2016deep}.
FL has been shown to be robust to unbalanced and non-IID data.

We use the \texttt{FederatedAveraging} algorithm presented
in~\citet{McMahan2017CommunicationEfficientLO} to aggregate client updates after
each round of local, on-device training to produce a new global model. At training round
$t$, a global model with parameters $w_t$, is sent to $K$ devices selected from
the device population. Each device has a local dataset $P_k$ which is split into
batches of size $B$. Stochastic gradient descent (SGD) is used on the
clients to compute new model parameters $w_{k}^{t+1}$. These client weights are
then averaged across devices, on the server, to compute the new model parameters $w_{t+1}$.

\section{Method}

\subsection{Network architecture}
The Long-Short-Term Memory (LSTM)~\cite{LSTM} architecture has been shown to
achieve state-of-the art performance of a number of sentiment prediction and
language modeling tasks~\cite{Radford2017LearningTG}.

We use an LSTM variant called the Coupled Input and Forget Gate
(CIFG)~\cite{Greff2017LSTMAS}. As with Gated Recurrent Units~\cite{GRU}, the
CIFG uses a single gate to control both the input and recurrent cell
self-connections. The input gate ($i$) and the forget gate ($f$) are related by $f = 1 - i$.
This coupling reduces the number of parameters per cell by 25\%, compared to an LSTM.

We use an input word vocabulary size of 10,000, an input embedding size of 96, and a
two-layer CIFG with 256 units per layer. The logits are passed through a softmax
layer to predict probabilities over 100 emoji.

\subsection{Pretraining}
~\citet{howard2018universal} demonstrated that pretraining parameters on a language modeling
task can improve performance on other tasks.

\begin{figure}
  \includegraphics[width=\columnwidth]{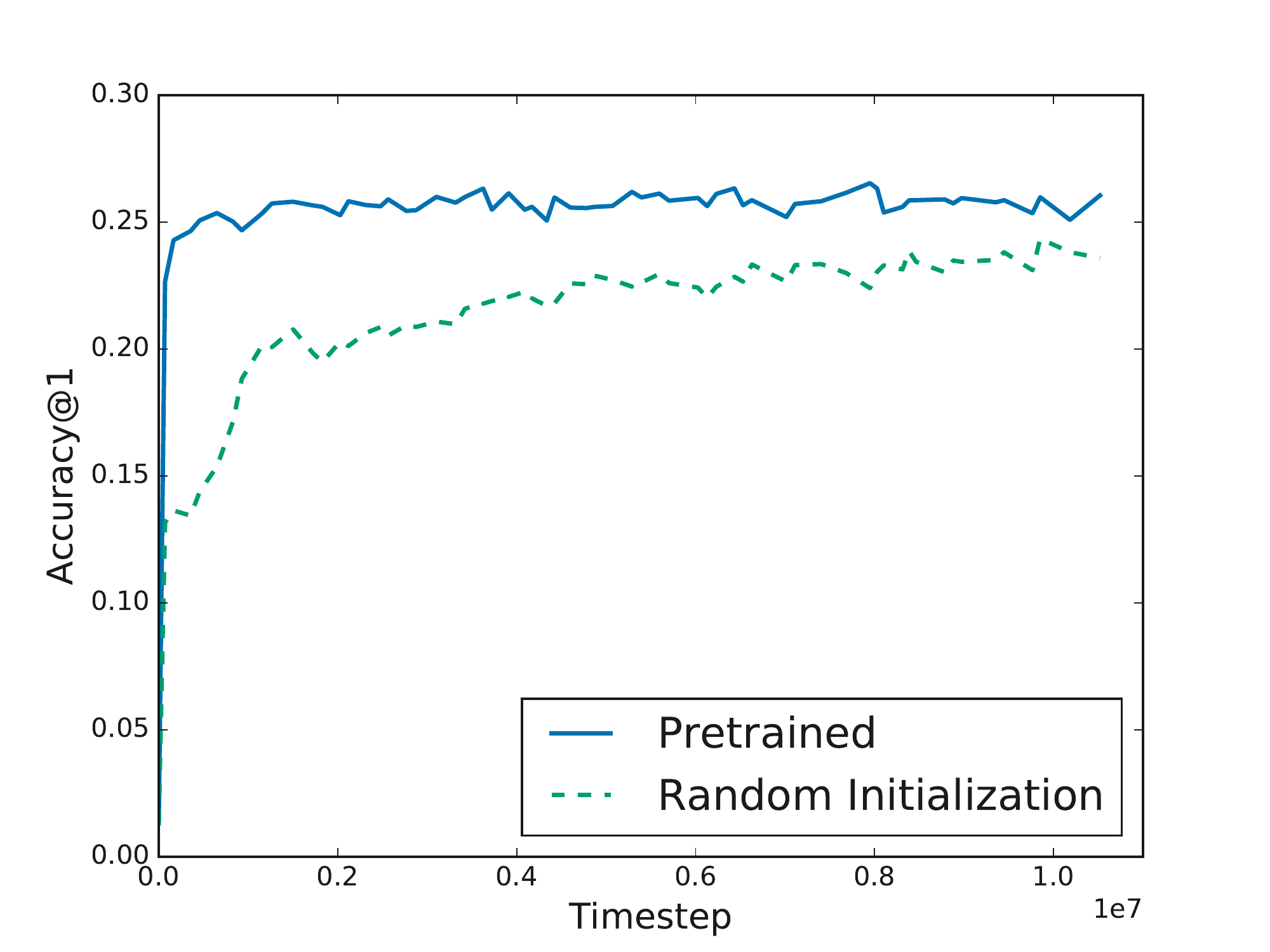}
  \caption{Accuracy@1 vs. training step with and without
           pretraining, using server-based evaluations.}
~\label{fig:pretraining}
\end{figure}

We pretrain all layers except the output projection layer, using a language
model trained to predict the next word in a sentence. For the output projection,
we reuse the input embeddings. This type of sharing of input and output
embeddings has been shown to improve performance of language
models~\cite{TiedEmbeddings}. Pretraining is done with federated learning using techniques similar to those
described by~\citet{hard2018federated}. The language model achieves an Accuracy@1
of 13.7\%, on the same vocabulary. Pretraining with a language model task leads to
much faster convergence for the emoji model, as seen in Figure~\ref{fig:pretraining}.

\subsection{Triggering}

In addition to predicting the correct emoji, a triggering mechanism must
determine when to show emoji predictions to users. For instance, a user
is likely to type \emoji{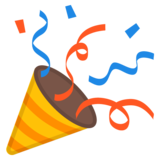} after typing ``Congrats'' or
``Congrats to you'' but not after ``Congrats to''.

One way to handle this would be to use a single language model that can predict
both words and emojis. However, we want to separate the task of predicting
relevant emoji from that of deciding how much we wanted emoji to trigger,
since the latter is more of a product decision, rather than a technical challenge.
For instance, if we want to allow users to control how often emoji predictions are offered,
it's easier to do with a separate model.

Another way to handle triggering is to use a separate binary classification model
that predicts the likelihood of the user typing any emoji after a given phrase.
However, using a separate  model for triggering 
leads to additional overhead in terms of memory and latency. Instead, we
adjust the softmax layer of the model to predict over $N$ emoji and an
additional unknown token \texttt{<UNK>} class. The \texttt{<UNK>} class is set
as the target output for inputs without emoji labels. At inference, we show the predictions from the model only if the probability of the \texttt{<UNK>}
class is less than a certain threshold.

During training, sentences without emoji are truncated to a random length in the
range [1, length of sentence]. Truncation allows the model to learn to not
predict emoji after conjunctions, prepositions etc. which typically occur in the
middle of sentences.

\subsection{Diversification}

\begin{figure}
  \includegraphics[width=\columnwidth]{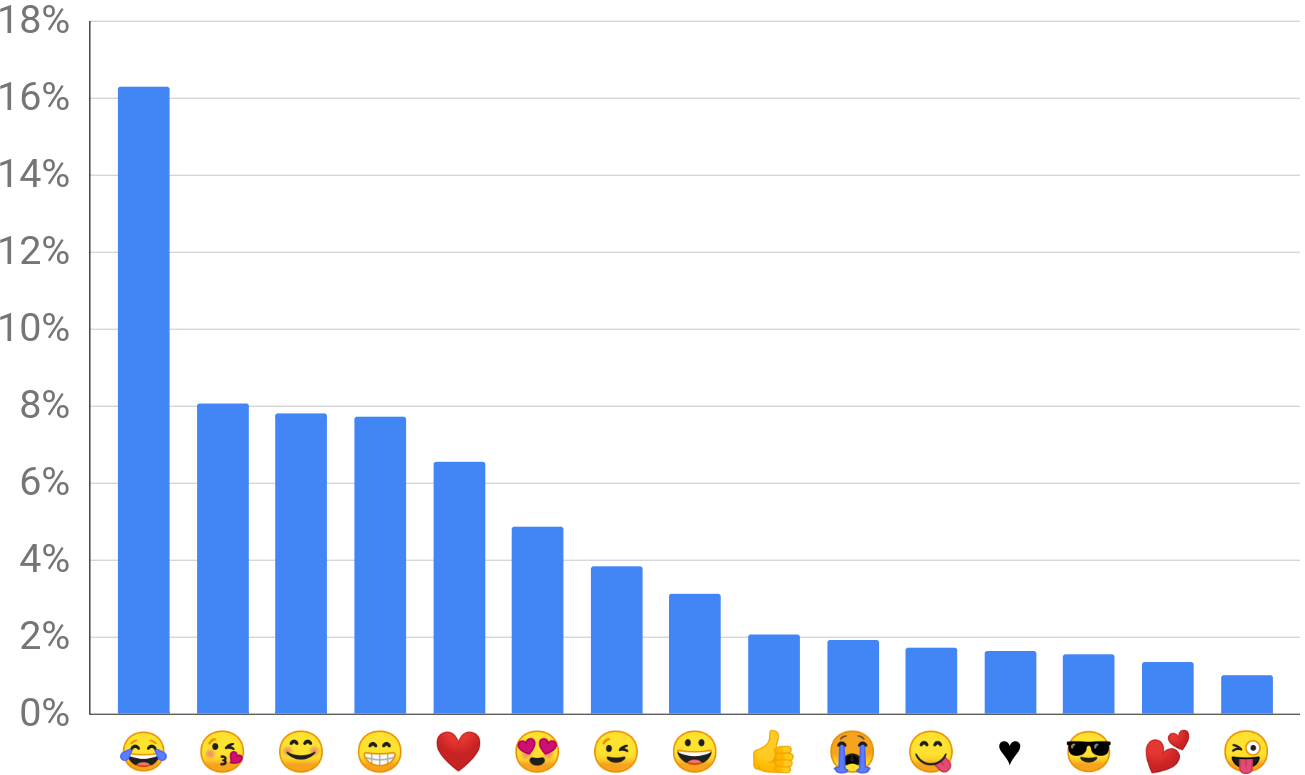}
  \caption{Distribution of 15 most frequently used emoji in English (US).}
  ~\label{fig:emoji-dist}
\end{figure}

The distribution of emoji usage frequency is very light-tailed as seen in
Figure~\ref{fig:emoji-dist}. As a result, the top predictions from the model are
almost always the most frequent emoji regardless of the input context. To
overcome this, the probability of each emoji($\widehat{P}$) is scaled by the
empirical probability of that emoji($P$) in the training data as follows.

\begin{equation}
S_i = \frac{\widehat{P} \left(\textrm{emoji}=i \vert \textrm{text} \right)}{{P\left(\textrm{emoji}=i\right)}^\alpha}
\end{equation}

where $\alpha$ is a scaling factor, determined empirically through experiments on live
traffic. Setting $\alpha$ to 0 removes diversification. Table~\ref{tab:table2}
provides examples with and without diversification.

\begin{table}[t!]
  \begin{tabular}{l c c}
    \toprule
    {\small\textit{Context}}
    & {\small \textit{$\alpha=0.0$}}
    & {\small \textit{$\alpha=0.7$}} \\
    \midrule
    Sorry I ended up falling asleep & \emoji{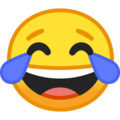} & \emoji{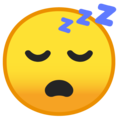} \\
    Good morning sunshine & \emoji{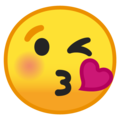} & \emoji{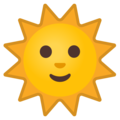} \\
    Coz I miss you xx & \emoji{kiss_face} & \emoji{kiss_face} \\
    I'm so sorry sweetie & \emoji{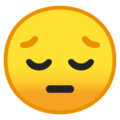}  & \emoji{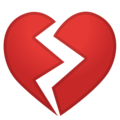} \\
    Hey girl you take it easy & \emoji{laughing_face} & \emoji{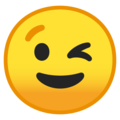} \\
    not sure what happened to that & \emoji{laughing_face} & \emoji{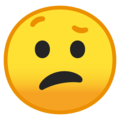} \\
    \bottomrule
  \end{tabular}
  \caption{Examples of emoji predictions with and without diversification}
  \label{tab:table2}
\end{table}

\section{Server-based Training}

Server-based training of models is done on data logged from Gboard users who have
opted to periodically share anonymized snippets of text typed in selected apps. All
personally identifiable information is stripped from these logs. The logs are
filtered further to only include sentences that are labeled as English with high
confidence by a language detection
model~\cite{Botha2017NaturalLP, Zhang2018AFC}. The subset of logs used for
training contain approximately 370 million snippets, approximately 11 million of
which contain emoji. Hyperparameters for server-based training are optimized using a
black-box optimization technique~\cite{vizier}.

\section{Federated Training}
The data used for federated training is stored in local caches on
client devices. For a device to participate in training, it must have at least
2 GB of RAM, must be located in United States or Canada, and must be using
English (US) as the primary language. In addition, only devices that
are connected to un-metered networks, idle, and charging are eligible for
participation at any given time. On average, each
client has approximately 400 sentences. The model is trained for one epoch on
each client, in each round. The model typically converges after 2000 training rounds.

In federated training, there is no explicit split of data into train and eval
samples. Instead, a separate evaluation task runs on a different subset of
client devices in parallel to the training task. The eval task uses model
checkpoints generated by the federated training task during a 24-hour period and
aggregates the metrics across evaluation rounds.

\section{Evaluation}

Model quality is evaluated using Accuracy@1, defined as the ratio of accurate
top-1 emoji predictions to the total number of examples containing emoji.
Area Under ROC Curve (AUC) is used to evaluate the quality of the triggering mechanism.
Computing the AUC involves numerical integration and is not straightforward to do in the FL setting.
Therefore, we report AUC only on logs data that is collected on the server. All
evaluation metrics are computed prior to diversification.

\section{Federated Experiments}

In FL, the contents of the client caches are constantly changing as old entries
are cleared and replaced by new activity. Since these experiments were conducted
non-concurrently, the client cache contents are different and therefore numbers
cannot be compared across experiments. We conduct experiments to study the
effect of client batch size ($B$), devices per round ($K$)  and server optimizer
configuration on model quality. We then take the best model and compare it with
a server trained model. The results are summarized in
Table~\ref{tab:table3}.

\begin{table}
  \centering
  \begin{tabular}{c l c c}
    \toprule
    {\small \textit{Experiment}} & {\small \textit{Accuracy@1}} & {\small \textit{AUC}} \\
    \midrule
    $B=1$   & 0.008 & 0.513 \\
    $B=10$  & 0.037 & 0.500 \\
    $B=50$  & 0.240 & 0.837 \\
    $B=200$ & 0.253 & 0.863 \\
    \midrule
    $K=20$  & 0.239 & 0.846 \\
    $K=50$  & 0.242 & 0.852 \\
    $K=200$ & 0.253 & 0.867 \\
    $K=500$ & 0.255 & 0.863 \\
    \midrule
    SGD, $\eta_s=1.0$ &  0.236 & 0.850 \\
    SGD, $\eta_s=2.0$ & 0.245 & 0.856 \\
    Momentum, $\eta_s=1.0$ & 0.247 & 0.856 \\
    \midrule
    Best federated & 0.256 & 0.863 \\
    Best server trained & 0.239 & 0.898 \\
    \bottomrule
  \end{tabular}
  \caption{The results from federated experiments. All numbers reported are
           after 2000 training rounds. $\eta_s$ refers to the learning rate used
           on the server for applying the update aggregated across users in each round.}
  \label{tab:table3}
\end{table}

Because of the sparsity of sentences containing emoji in the client caches, the
model quality is improved to a large degree by using large client batch sizes.
This is not entirely surprising, since gradient updates are more accurate with
larger batch sizes \cite{batchsize}. This is particularly true when the target
classes are heavily imbalanced.

The accuracy of the model also increases with the number of devices
per round but there are diminishing returns beyond $K=500$.

We experimented with various optimizers for the server update after each round
of federated training and found that using momentum of 0.9 with Nesterov
accelerated gradients~\cite{pmlr-v28-sutskever13} gives significant benefits
over using SGD, both in terms of speed of convergence and model performance.

\begin{figure}[t!]
  \includegraphics[width=\columnwidth]{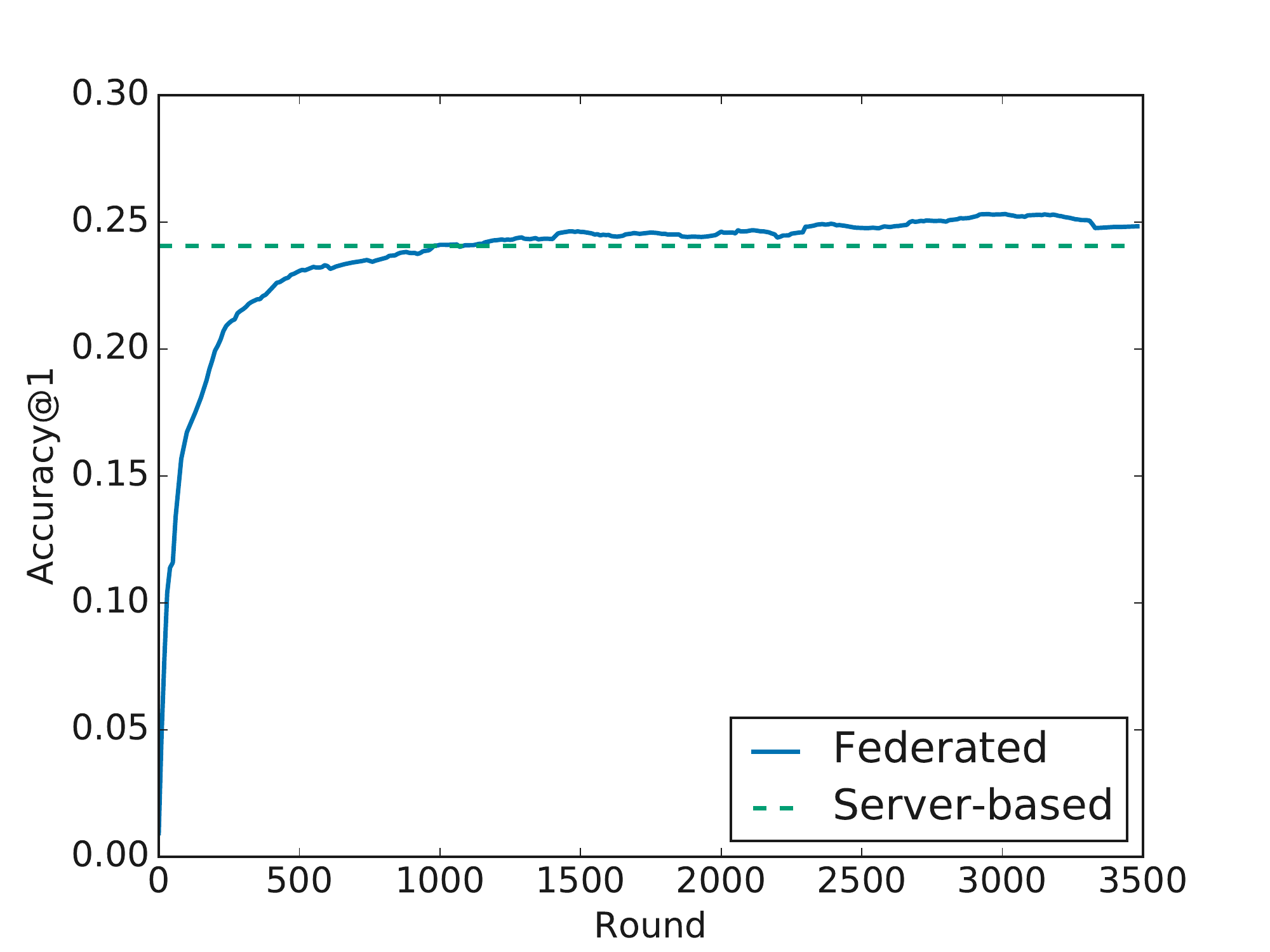}
\caption{Evaluation Accuracy@1 vs. Round for federated and server trained models.}
\label{tab:fed_vs_server}
\end{figure}

The best federated model, which runs in production, uses $B = 1000, K=1000$, and
is trained with momentum. We assign a weight of 0 to 99\% of the
\texttt{<UNK>} examples at training time so as to balance the triggering and
emoji prediction losses. We ran federated evaluation tasks of the best server-trained model on the
client caches in order to fairly compare the two training approaches. The
federated model achieved better Accuracy@1 in the federated evaluation, as shown in Figure \ref{tab:fed_vs_server}. However, the AUC achieved
by the federated model is lower than that of the server trained model.

AUC is only computed on the logs collected on the server.
These logs are restricted to short snippets
of text typed in selected apps, therefore the data is not believed to be as
representative of the text typed by users as data that resides on the client
caches. The lower AUC of the federated model is likely because of this bias.

\section{Live experiment}

At inference time, we use a quantized TensorFlow Lite~\cite{TFLite} model format. The
average inference latency is around 1 ms.

We ran a live-traffic experiment for users in USA and Canada typing in English
(US). We observed that both the federated and the server trained model lead to
significant increases in the overall click-through rate (CTR) of predictions,
total emoji shares, and daily active users (DAU) of emoji (see Table \ref{tab:table4}).
We also observed that the federated model did better than the server trained model on
all of the metrics.

Given that emoji are triggered rarely, the increase in CTR is quite large, for
both the models.

\newcommand\Tstrut{\rule{0pt}{2.6ex}} 

\begin{table}
  \centering
  \begin{tabular}{l c c}
    \toprule
    {\small\textit{Metric}}
    & \multicolumn{2}{c}{\small\textit{Relative change [\%]}}\\
    \midrule
    & \small\textit{Server trained} & \small\textit{Federated} \\
    \cline{2-3}
    Prediction CTR & $3.61 \pm 1.00$ & $3.66 \pm 0.95$  \Tstrut\\
    Emoji Shares & $3.63 \pm 0.99$ & $5.54 \pm 1.19$\\
    Emoji DAU & $9.57 \pm 0.39$ & $11.22 \pm 0.48$\\
    \bottomrule
  \end{tabular}
  \caption{Relative changes to metrics as a result of the server trained and
           federated emoji prediction models, measured in experiments on live user traffic.
           The baseline does not have any emoji predictions.
           Quoted 95\% confidence interval errors for all
           results are derived using the jackknife method with user buckets.}
  \label{tab:table4}
\end{table}

\section{Conclusions}

In this paper, we train an emoji prediction model using a CIFG-LSTM network. We
demonstrate that this model can be trained using FL to achieve
better performance than a server trained model. This work builds
on previous practical applications of federated learning
in~\citet{yang2018federated, hard2018federated, bonawitz2019towards}. We show that FL
works even with sparse data and poorly balanced classes.

\bibliography{acl2019}
\bibliographystyle{acl_natbib}

\end{document}